\documentclass[runningheads]{llncs}

\usepackage[T1]{fontenc}
\usepackage{amsmath}
\usepackage{graphicx,verbatim}
\usepackage{xcolor}
\usepackage{placeins}
\usepackage{array}
\usepackage{hyperref}

\usepackage{booktabs}

\begin{document}

\title{{AxonSynth}: Domain-Randomized Synthetic Data for Zero-Shot 3D Axon Segmentation in Light-Sheet Microscopy}
\titlerunning{{AxonSynth} for Zero-Shot Axon Segmentation}

\author{
Edward Gaibor\inst{1} \and
Kyriaki-Margarita Bintsi\inst{2} \and
Chiara Mauri\inst{2} \and
Emine {\"O}zen\inst{6} \and
Carmen Luz Leiva Ureta\inst{3} \and
Zayneb Bellatif\inst{4} \and
Chiara Maffei\inst{2} \and
Wenze Li\inst{5} \and
Zhuhao Wu\inst{8} \and
Elizabeth Hillman\inst{5,6} \and
Ya{\"e}l Balbastre\inst{7} \and
Anastasia Yendiki\inst{2}
}

\authorrunning{E. Gaibor et al.}

\institute{
Department of Computer Science, University of Massachusetts Boston,
Boston, MA, USA\\
\email{edward.gaibor001@umb.edu}
\and
Athinoula A. Martinos Center for Biomedical Imaging,
Massachusetts General Hospital and Harvard Medical School,
Charlestown, MA, USA
\and
Universidad San Sebasti{\'a}n, Chile
\and
Universit{\'e} Claude Bernard Lyon 1, Universit{\'e} de Lyon,
Lyon, France
\and
Department of Imaging Sciences, St. Jude Children's Research Hospital,
Memphis, TN, USA
\and
Department of Biomedical Engineering, Columbia University in the City of New York,
New York, NY, USA
\and
Department of Experimental Psychology, University College London,
London, United Kingdom
\and
Feil Family Brain and Mind Research Institute,
Weill Cornell Medicine, Cornell University,
New York, NY, USA
}

\maketitle

\begin{abstract}
Accurate segmentation  of axons in 3D microscopy data is important for analyzing white-matter organization, but dense ground truth labels are expensive to obtain. Existing supervised axon segmentation methods rely on target-domain annotations and can be brittle when tissue type, species, modality, or acquisition conditions change. We present AxonSynth, a domain-randomized synthetic-data framework for training 3D axon segmentation models without manually annotated real training volumes. AxonSynth generates dense synthetic axon labels with orientation priors that reflect realistic fiber configurations and renders them with randomized density, contrast, bias fields, blur, and noise. A three-class 3D U-Net is trained to predict background, axon sheath and intra-axonal space. We evaluate zero-shot transfer on 10 held-out light-sheet microscopy (LSM) patches from macaque and human brain samples labeled with one of three axonal markers, comparing against calibrated thresholding and Frangi filtering using overlap, corrected detection, false-positive, and topology metrics. On macaque samples, AxonSynth achieved the best corrected Dice and corrected precision (0.826 and 0.851), compared with 0.765 and 0.754 for thresholding and 0.685 and 0.762 for Frangi. On human samples, corrected Dice was comparable to thresholding (0.857 vs. 0.868), while component-count error decreased from 22,504 to 3,377. Across all held-out patches, AxonSynth reduced component-count error in 10/10 patches and Euler-characteristic error in 8/10. These results show that synthetic-label domain randomization can reduce dependence on manual axon annotation while supporting synthetic-to-real 3D segmentation.

\keywords{Data synthesis \and Domain randomization \and 3D axon segmentation \and Light sheet microscopy}
\end{abstract}

\section{Introduction}
% P1 — Motivation + annotation bottleneck
Accurate mapping of individual axons is essential for understanding neural circuits. Dense connectomics has therefore relied mainly on electron microscopy (EM), whose nanoscale resolution can resolve cells and synapses. Automated EM segmentation has advanced through supervised learning and large-scale reconstruction pipelines, including Google Connectomics flood-filling networks and MICrONS consortium petascale reconstruction methods~\cite{januszewski2018ffn,macrina2021petascale}. Axon-specific approaches such as AxonDeepSeg, DeepACSON, and multi-domain aggregation have addressed supervised axon and myelin segmentation across EM modalities and species~\cite{zaimi2018axondeepseg,abdollahzadeh2021deepacson,collin2024multidomain}. However, these axon-specific approaches remain dependent on labeled EM data or labeled examples from included target domains and can still require substantial proofreading.

% P2 — Prior work + gap
EM, however, does not scale to entire primate brains~\cite{shapsoncoe2024petavoxel}. Serial light-sheet microscopy (LSM) can cover much larger tissue volumes and, when combined with axonal-marker staining, offers a path toward microscale wiring diagrams at whole-primate-brain scale~\cite{park2024integrated}. Yet few off-the-shelf LSM axon segmentation models exist, and available methods often require target-domain adaptation or fine-tuning~\cite{friedmann2020mapping,li2023dlmbmap,oostrom2024trailmap}. Because manual annotation is costly, existing approaches do not directly address zero-shot 3D axon segmentation in LSM without target-domain labels.

% P3 — Synthetic data + DR as solution
Synthetic training data offers an alternative for this bottleneck. Domain randomization systematically alters contrast, noise, and other appearance properties during synthesis, which helps models learn robust structural features rather than modality-specific intensity and noise characteristics ~\cite{tobin2017domainrand}. This strategy has driven progress in brain MRI segmentation and synthetic-data-based vascular segmentation across imaging modalities~\cite{billot2023synthseg,chollet2024neurovascular,mauri2026vessynth}. These results motivate extending synthetic training to axons. Axons share some challenges with vessels, including small calibers and partial-volume effects, but they have directional coherence and form dense bundles, making separation of neighboring axons difficult. The key insight is that a model trained on sufficiently diverse synthetic data can generalize to real images not because the synthetic data are realistic, but because they span enough of the plausible appearance space.

% P4 — "Here, we present..."
Here, we present AxonSynth, a domain-randomized synthetic data pipeline that extends this strategy to 3D axon segmentation. The model predicts background, axon sheath (which may include the myelin and membrane of the axon), and intra-axonal space, and is evaluated against real LSM annotations by collapsing axon sheath and intra-axonal space into foreground. This paper makes two contributions: (1) a domain-randomized synthetic data pipeline for 3D axon segmentation with a three-class target definition, trained without manually annotated real data; and (2) quantitative zero-shot evaluation on annotated LSM patches spanning two species and three axonal markers. Code is available at \url{https://github.com/lincbrain/axonsynth}.

\begin{figure}[!htbp]
\centering
\includegraphics[width=\linewidth]{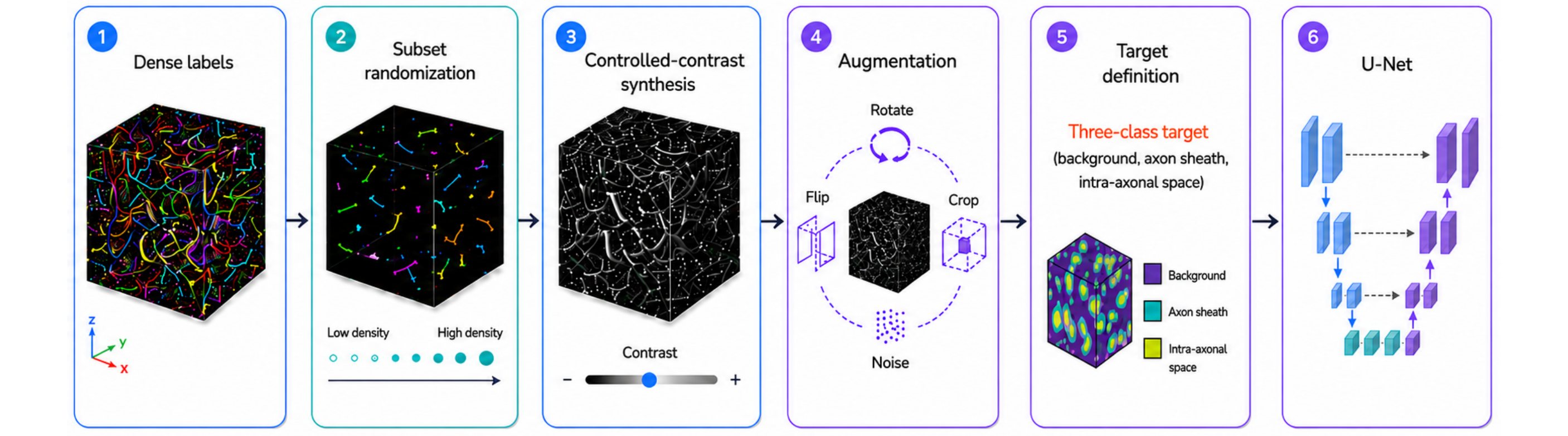}
\caption{AxonSynth training pipeline. Dense synthetic 3D axon instance labels are sampled as $128^3$ patches and converted into three-class supervision (background, axon sheath, intra-axonal space). A stochastic renderer applies domain randomization to generate diverse synthetic image--label pairs. A 3D U-Net is trained only on synthetic data; at inference on real LSM, axon sheath and intra-axonal space probabilities are summed to obtain the binary axon foreground.}
\label{fig:overview}
\end{figure}

\section{Methods}

\subsubsection{Synthetic labels.} 
% The training source consists of dense axon instance-label volumes and matching partial-volume probability maps. 
The synthetic label set contains 500 3D volumes, containing dense fiber configurations with $128{\times}128{\times}128$ voxels at 0.8\,$\mu$m isotropic resolution, split into 400 training and 100 validation volumes. Each source volume contains an object-ID map, where each axon has a separate integer ID, and a partial-volume probability map, which stores fractional axon occupancy per voxel. The object IDs are not segmentation targets, but allow random subsets of axons to be masked on-the-fly during training. Table~\ref{tab:synthetic-labels} summarizes the main geometry distributions. Compared with prior synthetic vessel pipelines for optical coherence tomography and multi-modal vessel segmentation~\cite{chollet2024neurovascular,mauri2026vessynth}, the main geometry change is the orientation prior: axons are sampled around a shared preferred direction to mimic dense bundles. The masked object-ID map can be converted into either a binary foreground target or a three-class background, axon sheath, and intra-axonal space target; the experiments below use the three-class target.

\subsubsection{Domain-randomized image synthesis.}
Each synthetic instance-label volume is converted into an image with a stochastic rendering pipeline. First, we generate a voxel-wise keep-probability field $p(x)$ to vary local axon density. The field family is sampled as linear, sigmoid, Gaussian, radial, or constant. For non-constant fields, lower and upper probability bounds are drawn from $U(0.05,0.4)$ and $U(0.6,1.0)$, respectively, and the field is scaled between these values. For constant fields, a single keep probability is drawn from $U(0.3,1.0)$. For each axon instance, we average $p(x)$ over all voxels belonging to that axon and keep the whole axon with that averaged probability. This selects complete axons while producing local density variation across the patch. Axon labels are then perturbed with spatially varying dilations and erosions.

Image intensities are generated separately for axons and background. Axons are grouped into random intensity classes, assigned intensities with a Gaussian mixture model (GMM), and rescaled to $[U(0.3,0.5),1.0]$. With probability 0.5, random smooth background label structures are sampled, eroded, assigned GMM intensities, rescaled to $[0,U(0.2,0.4)]$, and blended into non-axon voxels. Final image augmentations include a smooth additive bias field $b(x)$, applied voxel-wise as $I'(x)=I(x)+b(x)$ with values in approximately $[0,0.25]$, multiplicative bias fields, gamma correction sampled from $U(0,5)$, smoothing with width sampled from $U(0,2)$, chi/gamma noise, and quantile normalization. The objective is not photorealism. The renderer spans appearances beyond realistic microscopy while preserving axons being brighter than surrounding tissue, following the domain-randomization principle~\cite{chollet2024neurovascular}.

% TODO: Add a small synthetic sample?

\subsubsection{Segmentation targets.}
The three-class training target assigns each voxel to background, axon sheath, or intra-axonal space. Axon sheath voxels are identified by 3D face-adjacent random erosion of the synthetic foreground mask, and intra-axonal space voxels are the remaining foreground. For evaluation against binary real-data annotations, axon sheath and intra-axonal space probability maps are summed to produce a single foreground score. These classes are training targets only and do not imply a hollow axon morphology in the rendered image.

\begin{table}[t]
% [!htbp]
\centering
\fontsize{8}{9}\selectfont
\setlength{\tabcolsep}{3pt}
\caption{Synthetic label geometry. A tree denotes one synthetic axon instance before rasterization; the two-level maximum allows zero or one branch point. $U(a,b)$ denotes uniform sampling, and $\mathrm{LN}(\mu,\sigma)$ denotes LogNormal(mean, standard deviation).}
\label{tab:synthetic-labels}
\begin{tabular}{@{}>{\raggedright\arraybackslash}p{0.17\linewidth}>{\raggedright\arraybackslash}p{0.28\linewidth}>{\raggedright\arraybackslash}p{0.17\linewidth}>{\raggedright\arraybackslash}p{0.28\linewidth}@{}}
\toprule
Component & Setting & Component & Setting \\
\midrule
Label count & 500 volumes; 400 train, 100 validation & Patch and voxel & $128^3$ voxels; 0.8\,$\mu$m \\
Axon-tree count & $10^{U(3.13,3.63)}$ trees/patch ($\sim1350$--$4300$) & Orientation prior & Fixed $z$-axis Bingham; one component \\
Orientation variance & $\mathrm{LN}(0.1,0.05)$ & Radius & $\mathrm{LN}(10^{-3},5{\times}10^{-4})$ mm; clip $2{\times}10^{-3}$ mm \\
Radius change & $\mathrm{LN}(1,0.1)$ & Branch radius ratio & $\mathrm{LN}(1,0.1)$ \\
Branch count & $\mathrm{LN}(0.5,1)$ & Tortuosity & $\mathrm{LN}(1.1,2)$ \\
Levels per tree & Max. 2 (zero or one branch point) & Rasterization & B-spline curves; cosine partial volume \\
\bottomrule
\end{tabular}
\end{table}

% TODO: 3-class target diagram?

\subsubsection{Network and training.}
All experiments use a 3D U-Net~\cite{monai2022,cicek20163dunet} with encoder channel widths $\{16, 32, 64, 128, 256\}$, two residual units per level, dropout rate 0.1, and three softmax outputs. Dense axon labels are stored on disk, while image intensities are rendered from sampled label patches during training. The model uses a nested loss combining Dice and cross-entropy terms: one branch supervises foreground versus background and a second branch supervises axon sheath versus intra-axonal space, with the two terms summed with equal weight.

\section{Experiments}

\subsection {Dataset}
Swept-confocally aligned planar excitation (SCAPE) microscopy~\cite{ozen2025multicolor} is a single-objective light-sheet imaging method that enables submicron volumetric imaging of large brain sections with multiple 
simultaneous fluorescence channels. We use brain sections from human and macaque tissue, 
immunolabeled with neurofilament heavy chain (NEFH), neurofilament light chain (NEFL), 
and parvalbumin (PV) to target distinct axonal populations. The sections cover gray matter, 
white matter, and white--gray transition regions, providing a diverse set of axonal 
densities and configurations. Data were acquired at approximately $0.4\,\mu$m in-plane 
resolution. Due to the labor-intensive nature of manual annotation, expert-labeled ground 
truth is available for a limited subset: 12 patches with binary axon/background labels, 
annotated by an expert. Two patches (one human, one macaque) were reserved for threshold 
calibration and excluded from all held-out results; the remaining 10 patches (five human, 
five macaque) form the test set.

% \subsubsection{Light-sheet microscopy.}
% 12 annotated LSM patches from human and macaque tissue. Axonal markers are NEFH, NEFL, and PV. Includes gray matter, white matter, and white-gray transition regions. We have manual annota  tions as binary axon/background ground truth labels.

% \subsubsection{microCT.}
% We also evaluate qualitatively on one unlabeled microCT brain-tissue patch of $1217{\times}1036{\times}338$ voxels at $0.364\,\mu$m isotropic resolution, corresponding to approximately $0.44{\times}0.38{\times}0.12$ mm.

\subsection{Baselines.}
We compare AxonSynth with two classical baselines: raw-intensity thresholding and Frangi vesselness filtering~\cite{frangi1998multiscale}. For raw-intensity thresholding, we estimated one cutoff per domain on separate validation patches: 0.47 for human LSM and 0.46 for macaque LSM. These cutoffs were then fixed for evaluation, and the two validation patches were excluded from all held-out results.

For Frangi, vesselness was computed for bright ridges only, since axons appear as bright structures in LSM. Parameters were selected on the same validation patches by maximizing mean Dice score~\cite{dice1945measures}. Human LSM used $\alpha=0.25$, $\beta=0.5$, sigmas $\{3,4,5,6,7\}$, and threshold 0.05. Macaque LSM used $\alpha=0.25$, $\beta=0.33$, sigmas $\{2,3,4,5,6,7,8,9,10\}$, and threshold 0.01.

AxonSynth predicts background, axon sheath, and intra-axonal space. For evaluation, we sum the axon sheath and intra-axonal space probability maps and threshold the resulting foreground score. The calibrated foreground-score cutoffs were 0.94 for human LSM and 0.88 for macaque LSM.

\subsection{Evaluation metrics.}
Metrics are computed per patch after binarization and averaged. Let $P$, $G$, and $V$ denote the prediction, annotation, and valid voxel mask. Corrected metrics follow VesSynth~\cite{mauri2026vessynth} using a two-round 6-connected face-neighbor correction: starting from original true positives, adjacent false negatives are added to $P$ and adjacent false positives are removed from $P$, yielding $P_{\mathrm{corr}}$. Corrected counts are computed within $V$ as
\begin{equation}
\begin{aligned}
TP_{\mathrm{corr}} &= |P_{\mathrm{corr}}\cap G\cap V|, &
FP_{\mathrm{corr}} &= |P_{\mathrm{corr}}\cap \neg G\cap V|, \\
FN_{\mathrm{corr}} &= |\neg P_{\mathrm{corr}}\cap G\cap V|, &
TN_{\mathrm{corr}} &= |\neg P_{\mathrm{corr}}\cap \neg G\cap V|.
\end{aligned}
\end{equation}
We report corrected Dice, corrected precision (cPrec), corrected recall (cRec), and corrected false-positive rate (cFPR):
\begin{equation}
\begin{aligned}
\mathrm{Dice}_{\mathrm{corr}} &= \frac{2TP_{\mathrm{corr}}}{2TP_{\mathrm{corr}} + FP_{\mathrm{corr}} + FN_{\mathrm{corr}}}, &
\mathrm{cPrec} &= \frac{TP_{\mathrm{corr}}}{TP_{\mathrm{corr}} + FP_{\mathrm{corr}}}, \\
\mathrm{cRec} &= \frac{TP_{\mathrm{corr}}}{TP_{\mathrm{corr}} + FN_{\mathrm{corr}}}, &
\mathrm{cFPR} &= \frac{FP_{\mathrm{corr}}}{FP_{\mathrm{corr}} + TN_{\mathrm{corr}}}.
\end{aligned}
\end{equation}

Topology metrics are computed on the original binarized masks. We report $|\Delta\beta_0|$, where $\beta_0$ is the number of connected foreground components, and $|\Delta\chi|$, where $\chi=\beta_0-\beta_1+\beta_2$ is the Euler characteristic, $\beta_1$ is the number of tunnels, and $\beta_2$ is the number of enclosed cavities. Both $\Delta$ terms are absolute errors relative to the annotation.

\subsection{Training details.}
Training uses AdamW~\cite{loshchilov2019adamw}, mixed precision, and warmup followed by cosine learning-rate decay. Each cache draws 800 training samples and 200 validation samples from the synthetic label split in Table~\ref{tab:synthetic-labels}. Caches are reused for three epochs and then rebuilt, so the model sees 800 training sample presentations per epoch while the actual synthesized patches change every third epoch. The validation split is used only to select the best training epoch, and the selected checkpoint is used for all real-data evaluations.

\begin{table}[t]
% [!htbp]
\centering
\fontsize{8}{9}\selectfont
\setlength{\tabcolsep}{4pt}
\caption{Held-out LSM segmentation results on five human and five macaque annotated patches.
Calibrated thresholds (human vs. macaque): thresholding 0.47 vs. 0.46;
AxonSynth 0.94 vs. 0.88; Frangi 0.05 vs. 0.01.
$\uparrow$ higher is better; $\downarrow$ lower is better.}
\label{tab:lsm-results}
\begin{tabular}{lccccccc}
\hline
Method & Dice$\uparrow$ & Corr. Dice$\uparrow$ & cPrec.$\uparrow$ & cRec.$\uparrow$ & cFPR$\downarrow$ & $|\Delta\beta_0|$$\downarrow$ & $|\Delta\chi|$$\downarrow$ \\
\hline
\multicolumn{8}{l}{\textit{Human}} \\
Thresholding  & \textbf{0.576} & \textbf{0.868} & 0.910          & 0.837          & 0.0088          & 22{,}504          & 19{,}499         \\
AxonSynth     & 0.465          & 0.857          & 0.878          & \textbf{0.839} & 0.0109          & 3{,}377           & 3{,}692          \\
Frangi        & 0.512          & 0.769          & \textbf{0.970} & 0.639          & \textbf{0.0017} & \textbf{1{,}458}  & \textbf{1{,}006} \\
\midrule
\multicolumn{8}{l}{\textit{Macaque}} \\
Thresholding  & \textbf{0.550} & 0.765          & 0.754          & \textbf{0.852} & 0.0409          & 14{,}081          & 9{,}914          \\
AxonSynth     & 0.528          & \textbf{0.826} & \textbf{0.851} & 0.810          & \textbf{0.0141} & \textbf{808}      & \textbf{1{,}564} \\
Frangi        & 0.404          & 0.685          & 0.762          & 0.659          & 0.0211          & 1{,}216          & 2{,}557          \\
\hline
\end{tabular}
\end{table}

\section{Results}
Table~\ref{tab:lsm-results} reports performance on the 10 held-out LSM test patches, split into five human and five macaque patches across NEFH, NEFL, and PV axonal markers.

On human LSM, calibrated thresholding had the highest Dice and corrected Dice. AxonSynth had similar corrected Dice, 0.857 vs. 0.868, and similar recall, 0.839 vs. 0.837, but lower precision than thresholding. Bright-only Frangi had lower corrected Dice and recall, and its high precision and low cFPR/topology errors suggest under-segmentation.

On macaque LSM, AxonSynth had the highest corrected Dice and precision, 0.826 and 0.851, and lower cFPR than thresholding. Thresholding retained the highest Dice and recall but produced many more topological errors. Bright-only Frangi had lower corrected Dice and recall than both learned and thresholding-based predictions.

Topology metrics also differed by domain. In human LSM, AxonSynth reduced $|\Delta\beta_0|$ from 22{,}504 to 3{,}377 and $|\Delta\chi|$ from 19{,}499 to 3{,}692 relative to thresholding, while Frangi had the lowest topology errors but lower corrected Dice and recall. In macaque LSM, AxonSynth had the lowest topology errors, reducing $|\Delta\beta_0|$ from 14{,}081 to 808 and $|\Delta\chi|$ from 9{,}914 to 1{,}564 relative to thresholding. Across all 10 held-out patches, AxonSynth improved $|\Delta\beta_0|$ in all patches and $|\Delta\chi|$ in 8 patches. Thresholding predicted about 13.0 times the target component count, compared with 4.7 times for AxonSynth.

\begin{figure}[t]
% [!htbp]
\centering
\includegraphics[width=\linewidth]{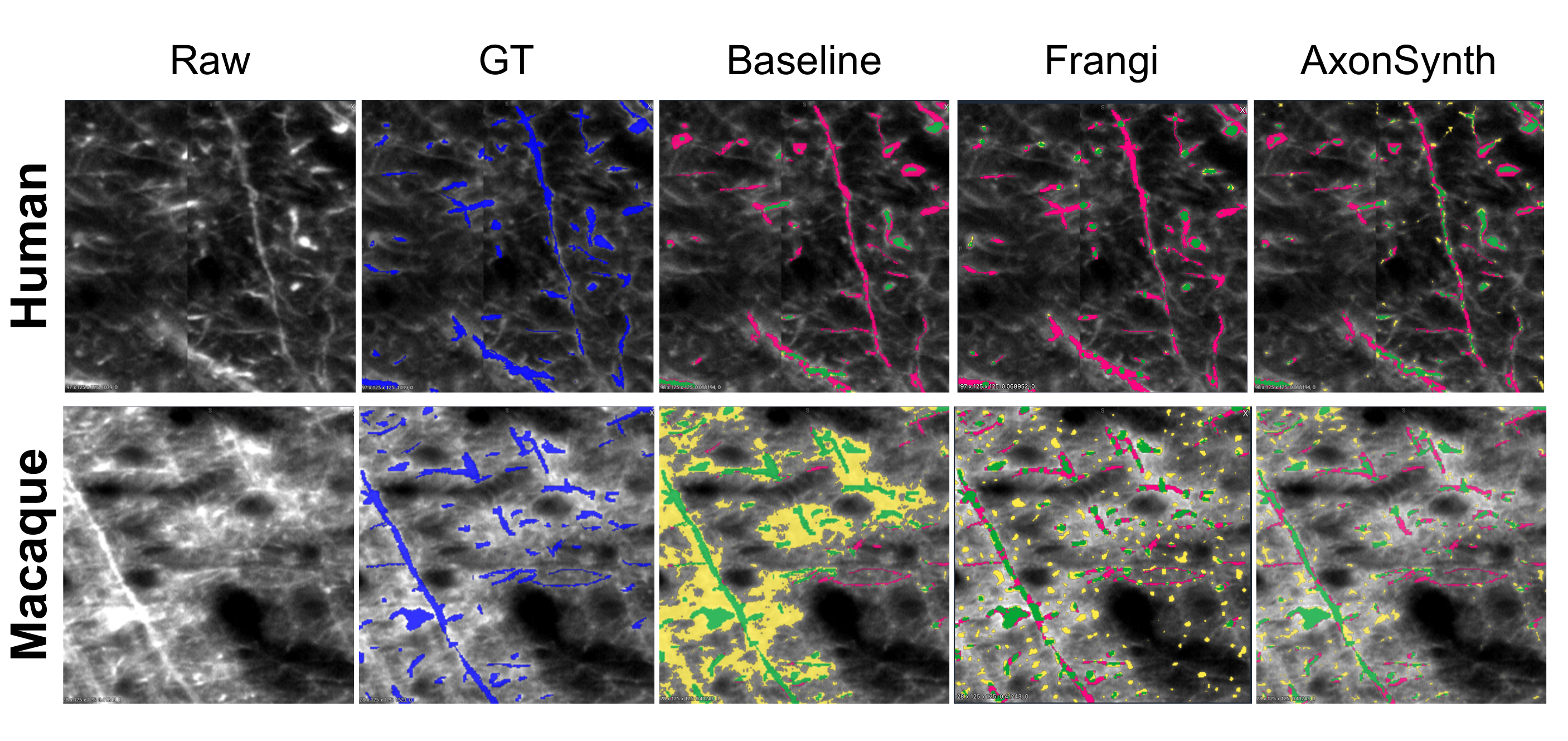}
\caption{Qualitative error comparison on two held-out LSM patches (human NEFH white matter; macaque NEFH gray matter). Ground truth (GT) axons are shown in blue. For each method, green denotes true positives, yellow false positives, and pink false negatives. Thresholding tends to over-segment bright background in macaque tissue, producing many fragmented false positives, whereas AxonSynth yields fewer spurious components. In the human example, AxonSynth recovers additional faint axons but introduces some extra false positives relative to thresholding.}
\label{fig:lsm-error-overlays}
\end{figure}

Figure~\ref{fig:lsm-error-overlays} shows different error patterns by domain. In the macaque patch, bright background causes thresholding to add many false-positive fragments. This matches Table~\ref{tab:lsm-results}: thresholding has the highest macaque Dice and recall, but lower corrected Dice, lower precision, and higher cFPR than AxonSynth. In the human patch, AxonSynth shows fewer false negatives, but the average human overlap metrics still favor thresholding. AxonSynth improves over thresholding in topology by lower $|\Delta\beta_0|$ and $|\Delta\chi|$ in both species, although Frangi has the lowest human topology errors.

\FloatBarrier

\subsection{Ablation and diagnostic analyses.}

\subsubsection{Two-class versus three-class targets.}
As a target ablation, we trained a two-class foreground/background model using the same synthetic label source and held-out LSM evaluation. The three-class axon sheath and intra-axonal space target improved corrected Dice from 0.785 to 0.842 and corrected precision from 0.742 to 0.865, while lowering cFPR from 0.0285 to 0.0125. The two-class model had higher recall, 0.863 versus 0.825, and lower mean topology errors ($|\Delta\beta_0|$ 1{,}852 versus 2{,}093; $|\Delta\chi|$ 1{,}369 versus 2{,}628), but its lower precision indicates more foreground oversegmentation. We therefore use the three-class model as the primary AxonSynth configuration.

\subsubsection{LSM depth analysis.}
Modern LSM systems, such as the one used in this study, use an oblique light-sheet, leading to a depth-dependent contrast. We therefore analyzed performance as a function of normalized imaging depth across all 12 annotated LSM patches. This diagnostic measures within-stack degradation rather than held-out model selection. Both thresholding and AxonSynth degraded with depth: corrected Dice slopes were $-0.135$ and $-0.141$ per unit normalized depth, respectively. The deepest slab had corrected Dice lower than the most superficial one by 0.124 for thresholding and 0.133 for AxonSynth. Mean normalized intensity also decreased with depth, with slope $-0.0566$, consistent with signal loss along the stack. The effect was stronger in macaque patches, with corrected Dice slopes of $-0.215$ for thresholding and $-0.214$ for AxonSynth, than in human patches, with slopes of $-0.055$ and $-0.068$.

\section{Discussion \& Conclusion}

AxonSynth shows that synthetic axon labels with broad appearance randomization can train a 3D segmentation model with transfer to real LSM without real training labels. When compared to thresholding, AxonSynth's advantage is not uniformly higher overlap, but rather lower fragmented components and Euler-characteristic error. 

The target ablation indicates that separating axon sheath and intra-axonal space during training improves real-data precision and corrected Dice. This may seem paradoxical as the axonal markers used here do not label myelin sheaths. Empirically, however, the three-class model appears to address the main failure mode in dense axon bundles: neighboring axons can be merged when the model sees only a binary foreground target. The two-class model had higher recall and lower mean topology errors, but its lower precision and higher cFPR indicate more foreground oversegmentation. The three-class target is therefore the better default for conservative axon detection, although the best target definition may depend on whether the downstream task prioritizes recall, topology, or instance separation.

\begin{comment}
Limitations and future work include broader validation on additional labeled LSM patches and comparison with 3D networks trained on real annotations. The synthetic generator is intentionally simplified and does not aim to produce realistic axon anatomy or photorealistic microscopy. Future work will improve topology correction to further preserve axon continuity, specialize synthesis for human patches, and model depth-dependent signal attenuation.
\end{comment}

Depth-wise degradation suggests adding synthesis that models attenuation and blurring to achieve more consistent performance across imaging depths. Other future work includes broader LSM validation, testing in other microscopy modalities, and comparing with 3D networks trained on real annotations.

Overall, AxonSynth is a promising first step toward automatic large-scale 3D axon segmentation from synthetic training data. It supports zero-shot segmentation on real human and macaque LSM, reduces topology errors relative to thresholding, and maintains balanced detection across species.

\begin{credits}
\subsubsection{\ackname} 
This work was supported by the center for Large-scale Imaging of Neural Circuits (LINC), an NIH BRAIN Initiative Connectivity across Scales (CONNECTS) comprehensive center (UM1-NS132358). Additional support was provided by the National Institute for Mental Health (R01-MH045573, P50-MH106435) and the National Institute for Neurological Disorders and Stroke (R01-NS119911, R01-NS127353).

\subsubsection{\discintname}
The authors have no competing interests to declare that are
relevant to the content of this article. 
\end{credits}

\bibliographystyle{splncs04}
\bibliography{refs}

@article{chollet2024neurovascular,
  title = {Neurovascular Segmentation in {sOCT} with Deep Learning and Synthetic Training Data},
  author = {Chollet, Etienne and Balbastre, Ya{\"e}l and Mauri, Chiara and Magnain, Caroline and Fischl, Bruce and Wang, Hui},
  journal = {arXiv preprint arXiv:2407.01419},
  year = {2024},
  url = {https://arxiv.org/abs/2407.01419}
}

@inproceedings{cicek20163dunet,
  author = {{\c{C}}i{\c{c}}ek, {\"O}zg{\"u}n and Abdulkadir, Ahmed and Lienkamp, Soeren S. and Brox, Thomas and Ronneberger, Olaf},
  title = {{3D U-Net}: Learning Dense Volumetric Segmentation from Sparse Annotation},
  booktitle = {Medical Image Computing and Computer-Assisted Intervention},
  pages = {424--432},
  year = {2016},
  publisher = {Springer}
}

@inproceedings{tobin2017domainrand,
  author = {Tobin, Josh and Fong, Rachel and Ray, Alex and Schneider, Jonas and Zaremba, Wojciech and Abbeel, Pieter},
  title = {Domain Randomization for Transferring Deep Neural Networks from Simulation to the Real World},
  booktitle = {IEEE/RSJ International Conference on Intelligent Robots and Systems},
  pages = {23--30},
  year = {2017},
  publisher = {IEEE}
}

@article{billot2023synthseg,
  author = {Billot, Benjamin and Greve, Douglas N. and Puonti, Oula and Thielscher, Axel and Van Leemput, Koen and Fischl, Bruce and Dalca, Adrian V. and Iglesias, Juan Eugenio},
  title = {{SynthSeg}: Segmentation of Brain {MRI} Scans of Any Contrast and Resolution Without Retraining},
  journal = {Medical Image Analysis},
  volume = {86},
  pages = {102789},
  year = {2023}
}

@inproceedings{loshchilov2019adamw,
  author = {Loshchilov, Ilya and Hutter, Frank},
  title = {Decoupled Weight Decay Regularization},
  booktitle = {International Conference on Learning Representations},
  year = {2019}
}

@article{monai2022,
  author = {Cardoso, M. Jorge and Li, Wenqi and Brown, Richard and Ma, Nic and Kerfoot, Eric and Wang, Yiheng and Murray, Benjamin and Myronenko, Andriy and Zhao, Can and Yang, Dong and others},
  title = {{MONAI}: An Open-Source Framework for Deep Learning in Healthcare},
  journal = {arXiv preprint arXiv:2211.02701},
  year = {2022},
  doi = {10.48550/arXiv.2211.02701}
}

@article{mauri2026vessynth,
  author = {Mauri, Chiara and McKenzie, A. and Analoro, C. and Yeon, E. and Coviello, R. and Mora, J. and Chollet, Etienne and Binder, L. D. and Mahar, A. and Lin, S. and Benlahcen, M. and Ream, A. and Jama, A. and Garcia, I. and Tran, N. and Onta, P. and Wood, S. and Willis, A. and Mahmood, A. and Sinoballa, G. and Malki, A. and Tran, K. and Malireddy, V. and Onumajuru, N. and Lakshmanan, S. and Landaverde, K. H. and Sidow, R. and Wood, D. and Nguyen, B. and Hernandez, J. and Bernier, M. and Hunter, J. and Tum, A. and Chavez, V. and Shahu, Z. and Vasi, I. and Visser, A. and Ghaouta, Z. and Bond, F. and Vigneshwaran, R. and Kirkpatrick, E. and Barbosa, M. A. and Rauh, K. and Herisse, R. and Pallares, E. G. and Zeng, X. and Varadarajan, D. and Wang, Hui and Magnain, Caroline and Edlow, B. L. and Hoffmann, M. and Fischl, Bruce and Balbastre, Ya{\"e}l},
  title = {{VesSynth}: Tubes Are All You Need for Robust Cross-Scale Cross-Modal {3D} Vessel Segmentation},
  journal = {bioRxiv},
  year = {2026},
  pages = {2026.04.01.715909},
  doi = {10.64898/2026.04.01.715909},
  pmid = {41959422},
  pmcid = {PMC13060244},
  url = {https://pmc.ncbi.nlm.nih.gov/articles/PMC13060244/}
}

@article{zaimi2018axondeepseg,
  author = {Zaimi, Aldo and Wabartha, Maxime and Herman, Victor and Antonsanti, Pierre-Louis and Perone, Christian S. and Cohen-Adad, Julien},
  title = {{AxonDeepSeg}: Automatic Axon and Myelin Segmentation from Microscopy Data Using Convolutional Neural Networks},
  journal = {Scientific Reports},
  volume = {8},
  pages = {3816},
  year = {2018},
  doi = {10.1038/s41598-018-22181-4}
}

@article{collin2024multidomain,
  author = {Collin, Armand and Boschet, Arthur and Boudreau, Mathieu and Cohen-Adad, Julien},
  title = {Multi-Domain Data Aggregation for Axon and Myelin Segmentation in Histology Images},
  journal = {arXiv preprint arXiv:2409.11552},
  year = {2024},
  doi = {10.48550/arXiv.2409.11552}
}

@article{abdollahzadeh2021deepacson,
  author = {Abdollahzadeh, Ali and Belevich, Ilya and Jokitalo, Eija and Sierra, Alejandra and Tohka, Jussi},
  title = {{DeepACSON} Automated Segmentation of White Matter in {3D} Electron Microscopy},
  journal = {Communications Biology},
  volume = {4},
  pages = {179},
  year = {2021},
  doi = {10.1038/s42003-021-01699-w}
}

@article{januszewski2018ffn,
  author = {Januszewski, Micha{\l} and Kornfeld, J{\"o}rgen and Li, Peter H. and Pope, Art and Blakely, Tim and Lindsey, Larry and Maitin-Shepard, Jeremy B. and Tyka, Mike and Denk, Winfried and Jain, Viren},
  title = {High-Precision Automated Reconstruction of Neurons with Flood-Filling Networks},
  journal = {Nature Methods},
  volume = {15},
  number = {8},
  pages = {605--610},
  year = {2018},
  doi = {10.1038/s41592-018-0049-4}
}

@article{macrina2021petascale,
  author = {Macrina, Thomas and Lee, Kisuk and Lu, Ran and Turner, Nicholas L. and Wu, Jingpeng and others},
  title = {Petascale Neural Circuit Reconstruction: Automated Methods},
  journal = {bioRxiv},
  year = {2021},
  doi = {10.1101/2021.08.04.455162},
  note = {Preprint}
}

@inproceedings{frangi1998multiscale,
  author = {Frangi, Alejandro F. and Niessen, Wiro J. and Vincken, Koen L. and Viergever, Max A.},
  title = {Multiscale Vessel Enhancement Filtering},
  booktitle = {Medical Image Computing and Computer-Assisted Intervention},
  pages = {130--137},
  year = {1998},
  publisher = {Springer}
}

@article{dice1945measures,
  author = {Dice, Lee R.},
  title = {Measures of the Amount of Ecologic Association Between Species},
  journal = {Ecology},
  volume = {26},
  number = {3},
  pages = {297--302},
  year = {1945},
  doi = {10.2307/1932409}
}

@inproceedings{ozen2025multicolor,
  title={Multicolor High Resolution SCAPE microscopy for Understanding Neural Connectivity},
  author={{\"O}zen, Emine and Yan, Richard W and Li, Wenze and Evans, Emily and Casper, Malte J and Wang, Wei and Bell, Elissa and Chai, Kaidong and Shao, Jasmine and Wu, Jingjing and others},
  booktitle={Optics and the Brain},
  pages={BW1B--3},
  year={2025},
  organization={Optica Publishing Group}
}

@article{shapsoncoe2024petavoxel,
  author = {Shapson-Coe, Alexander and Januszewski, Micha{\l} and Berger, Daniel R. and Pope, Art and Wu, Yuelong and Blakely, Tim and Schalek, Richard and Li, Peter H. and Wang, Shuohong and Maitin-Shepard, Jeremy and Karlupia, Neha and Dorkenwald, Sven and Sj{\"o}stedt, Evelina and Leavitt, Laramie and Lee, Dongil and Troidl, Jakob and Collman, Forrest and Bailey, Luke and Fitzmaurice, Angerica and Kar, Rohin and Field, Benjamin and Wu, Hank and Wagner-Carena, Julian and Aley, David and Lau, Joanna and Lin, Zudi and Wei, Donglai and Pfister, Hanspeter and Peleg, Adi and Jain, Viren and Lichtman, Jeff W.},
  title = {A petavoxel fragment of human cerebral cortex reconstructed at nanoscale resolution},
  journal = {Science},
  year = {2024},
  volume = {384},
  number = {6696},
  pages = {eadk4858},
  doi = {10.1126/science.adk4858}
}

@article{park2024integrated,
  author = {Park, Juhyuk and Wang, Ji and Guan, Webster and Gjesteby, Lars and Pollack, Dylan and Kamentsky, Lee and Evans, Nicholas B. and Stirman, Jeff and Gu, Xinyi and Zhao, Chuanxi and Marx, Slayton and Kim, Minyoung E. and Choi, Seo Woo and Snyder, Michael P. and Chavez, David and Su-Arcaro, Clover and Tian, Yuxuan and Park, Chang Sin and Zhang, Qiangge and Yun, Dae Hee and Moukheiber, Mira and Feng, Guoping and Yang, X. William and Keene, C. Dirk and Hof, Patrick R. and Ghosh, Satrajit and Frosch, Matthew P. and Brattain, Laura J. and Chung, Kwanghun},
  title = {Integrated platform for multiscale molecular imaging and phenotyping of the human brain},
  journal = {Science},
  year = {2024},
  volume = {384},
  number = {6701},
  pages = {eadh9979},
  doi = {10.1126/science.adh9979}
}

@article{friedmann2020mapping,
  author = {Friedmann, Drew and Pun, Albert and Adams, Eliza L. and Lui, Jan H. and Kebschull, Justus M. and Grutzner, Sophie M. and Castagnola, Caitlin and Tessier-Lavigne, Marc and Luo, Liqun},
  title = {Mapping mesoscale axonal projections in the mouse brain using a 3D convolutional network},
  journal = {Proceedings of the National Academy of Sciences of the United States of America},
  year = {2020},
  volume = {117},
  number = {20},
  pages = {11068--11075},
  doi = {10.1073/pnas.1918465117}
}

@article{li2023dlmbmap,
  author = {Li, Zhongyu and Shang, Zengyi and Liu, Jingyi and Zhen, Haotian and Zhu, Entao and Zhong, Shilin and Sturgess, Robyn N. and Zhou, Yitian and Hu, Xuemeng and Zhao, Xingyue and Wu, Yi and Li, Peiqi and Lin, Rui and Ren, Jing},
  title = {{D-LMBmap}: a fully automated deep-learning pipeline for whole-brain profiling of neural circuitry},
  journal = {Nature Methods},
  year = {2023},
  volume = {20},
  number = {10},
  pages = {1593--1604},
  doi = {10.1038/s41592-023-01998-6}
}

@article{oostrom2024trailmap,
  author = {Oostrom, Marjolein and Muniak, Michael A. and Eichler West, Rogene M. and Akers, Sarah and Pande, Paritosh and Obiri, Moses and Wang, Wei and Bowyer, Kasey and Wu, Zhuhao and Bramer, Lisa M. and Mao, Tianyi and Webb-Robertson, Bobbie Jo M.},
  title = {Fine-tuning {TrailMap}: The utility of transfer learning to improve the performance of deep learning in axon segmentation of light-sheet microscopy images},
  journal = {PLOS ONE},
  year = {2024},
  volume = {19},
  number = {3},
  pages = {e0293856},
  doi = {10.1371/journal.pone.0293856}
}

\end{document}